\documentclass[10pt,twocolumn,letterpaper]{article}

\usepackage{cvpr}
\usepackage{times}
\usepackage{epsfig}
\usepackage{graphicx}
\usepackage{amsmath}
\usepackage{amssymb}
\usepackage{tabularx,ragged2e}

\usepackage{booktabs,calc}
\usepackage{multirow}
\usepackage{hhline}
\usepackage{array}
\usepackage{float}
\usepackage{makecell}
\usepackage{xcolor}
\usepackage{mathtools}
\usepackage{enumitem}
\usepackage{cite}
\usepackage{dsfont}
\usepackage{pifont}
\usepackage{balance}

\newcolumntype{Y}{>{\centering\arraybackslash}X}
\newcolumntype{s}{>{\hsize=.3\hsize}Y}
\newcolumntype{t}{>{\hsize=.2\hsize}Y}
\DeclareMathOperator*{\argmin}{argmin}

\usepackage[pagebackref=true,breaklinks=true,letterpaper=true,colorlinks,bookmarks=false]{hyperref}

\cvprfinalcopy 

\def\cvprPaperID{000} 

\begin{document}
\title{Weakly-Supervised 3D Human Pose Learning via Multi-view Images in the Wild}

\author{Umar Iqbal~~~~~~~Pavlo Molchanov~~~~~~~Jan Kautz\\
NVIDIA \\
{\tt\small\{uiqbal,\,pmolchanov,\,jkautz\}@nvidia.com}
}

\maketitle


\newcommand{\UI}[1]{{\color{pink}UI: #1}}
\newcommand{\pmnote}[1]{{\color{orange}PM: #1}}
\newcommand{\OH}[1]{{\color{blue}[OH: #1]}}
\newcommand{\JK}[1]{{\color{cyan}[JK: #1]}}

\newcommand{\figref}[1]{Fig.~\ref{#1}}
\newcommand{\tabref}[1]{Table ~\ref{#1}}

\mathchardef\mhyphen="2D

\begin{abstract}
One major challenge for monocular 3D human pose estimation in-the-wild is the acquisition of training data that contains unconstrained images annotated with accurate 3D poses. In this paper, we address this challenge by proposing a weakly-supervised approach that does not require 3D annotations and learns to estimate 3D poses from unlabeled multi-view data, which can be acquired easily in  in-the-wild  environments. We propose a novel end-to-end learning framework that enables weakly-supervised training using multi-view consistency. Since multi-view consistency is prone to degenerated solutions, we adopt a 2.5D pose representation and propose a novel objective function that can only be minimized when the predictions of the trained model are consistent and plausible across all camera views. We evaluate our proposed approach on two large scale datasets (Human3.6M and MPII-INF-3DHP) where it achieves state-of-the-art performance among semi-/weakly-supervised methods. 
\vspace{-5mm}
\end{abstract}

\section{Introduction}

Learning to estimate 3D body pose from a single RGB image is of great interest for many practical applications. The state-of-the-art methods~\cite{LiC14,Sijin2015iccv,zhou2016deep,tekin2016structured,popa2017CVPRmultitask,pavlakos2017volumetric,sun2017compositional,zhou2017weakly, dabral18SFM, sun18integeral} in this area use images annotated with 3D poses and  train deep neural networks to directly regress 3D pose from images. While the performance of these methods has improved significantly, their applicability in in-the-wild environments has been limited due to the lack of training data with ample diversity. The commonly used training datasets such as Human3.6M~\cite{h36m_pami}, and MPII-INF-3DHP~\cite{mono20173dhp} are collected in controlled indoor settings using sophisticated multi-camera motion capture systems. While scaling such systems to unconstrained outdoor environments is impractical, manual annotations are difficult to obtain and prone to errors. Therefore, current methods resort to existing training data and try to improve the generalizabilty of trained models by incorporating additional weak supervision in form of various 2D annotations for in-the-wild images~\cite{zhou2017weakly, sun2017compositional, pavlakos2018ordinal}. While 2D annotations can be obtained easily, they do not provide sufficient information about the 3D body pose, especially when the body joints are foreshortened or occluded. Therefore, these methods rely heavily on the ground-truth 3D annotations, in particular, for depth predictions.  

Instead of using 3D annotations, in this work, we propose to use unlabeled multi-view data for training. We assume this data to be without extrinsic camera calibration. Hence, it can be collected very easily in any in-the-wild setting. In contrast to 2D annotations, using multi-view data for training has several obvious advantages \eg, ambiguities arising due to body joint occlusions as well as foreshortening or motion blur can be resolved by utilizing information from other views. There have been only few works~\cite{pavlakos2017harvesting, rohdin2018multiview, rohdin2018geometry, kocabas2019epipolar} that utilize multi-view data to learn monocular 3D pose estimation models. While the approaches~\cite{pavlakos2017harvesting,rohdin2018geometry} need extrinsic camera calibration,~\cite{rohdin2018geometry,rohdin2018multiview} require at least some part of their training data to be labelled with ground-truth 3D poses. Both of these requirements are, however, very hard to acquire for unconstrained data, hence, limit the applicability of these methods to controlled indoor settings.  In~\cite{kocabas2019epipolar}, 2D poses obtained from multiple camera views are used to generate pseudo ground-truths for training. However, this method uses a pre-trained pose estimation model which remains fixed during training, meaning 2D pose errors remain unaddressed and can propagate to the generated pseudo ground-truths. 

\begin{figure*}[t]
\includegraphics[trim={0.0cm .15cm 0.0cm 0},clip,scale=1]{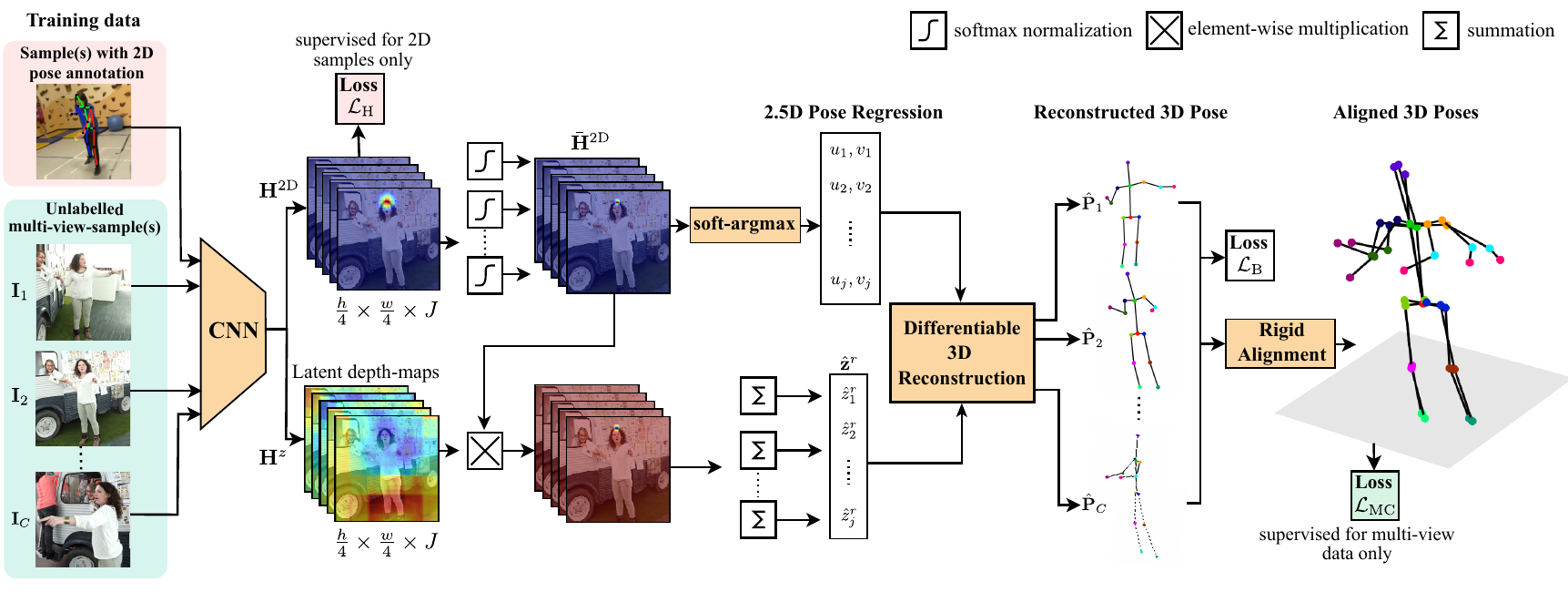}
\caption{An end-to-end approach for learning 3D pose estimation model without 3D annotations. For training, we only use unlabeled multi-view data along with an independent collection of images with 2D pose annotations. Given an RGB image, the model is trained to generate 2D heatmaps $\mathbf{H}^{\mathrm{2D}}$ and latent depth-maps $\mathbf{H}^{z}$ - shown only for $\mathbf{I}_2$ for simplicity. The 2D heatmaps are converted to 2D pose coordinates using $\mathrm{soft{\mhyphen}argmax}$. The relative-depth values $\mathbf{\hat{z}}^r$ are obtained by taking channel-wise summation of the multiplication of normalized heatmaps $\bar{\mathbf{H}}^{\mathrm{2D}}$ and latent depth-maps $\mathbf{H}^{z}$. The 3D pose is reconstructed in a fully differentiable manner by exploiting the scale normalization constraint (Sec.~\ref{sec:2.5D_pose}). The images with 2D pose annotation are used for heatmap loss $\mathcal{L}_\text{H}$. The 3D supervision is provided via a multi-view consistency loss $\mathcal{L}_{\text{MC}}$ that enforces that the 3D poses generated from different views should be identical up to a rigid transform. Given 2D pose estimates from different views and camera intrinsics,  the objective is designed such that the only way for the network to minimize it is to produce correct relative depth values $\hat{\mathbf{z}}^r$ (Sec.~\ref{sec:weakly-supervised-training}). We also enforce a bone-length loss $\mathcal{L}_\text{B}$  on each predicted 3D pose to further constrain the search space.\vspace{-5mm}} 
\label{fig:overview}
\end{figure*}

In this work, we present a weakly-supervised approach for monocular 3D pose estimation that does not require any 3D pose annotations at all. For training, we only use a collection of unlabeled multi-view data and an independent collection of images annotated with 2D poses. An overview of the approach can be seen in Fig.~\ref{fig:overview}. Given an RGB image as input, we train the network to predict a 2.5D pose representation~\cite{iqbal2018hand} from which the 3D pose can be reconstructed in a fully-differentiable way. Given unlabeled multi-view data, we use a multi-view consistency loss which enforces the 3D poses estimated from different views to be consistent up to a rigid transformation. However, naively enforcing multi-view consistency can lead to degenerated solutions. We, therefore, propose a novel objective function which is constrained such that it can only be minimized when the 3D poses are predicted correctly from all camera views.  The proposed approach can be trained in a fully end-to-end manner, it does not require extrinsic camera calibration and is robust to body part occlusions and truncations in the unlabeled multi-view data. Furthermore, it can also improve the 2D pose predictions by exploiting multi-view consistency during training.

We evaluate our approach on two large scale datasets where it outperforms existing methods for semi-/weakly-supervised methods by a large margin. We also show that  the MannequinChallenge dataset~\cite{li2019mannequin}, which provides in-the-wild videos of people in static poses, can be effectively exploited by our proposed method to improve the generalizability of trained models, in particular, when their is a significant domain gap between the training and testing environments.

\section{Related Work}

We discuss existing methods for monocular 3D human pose estimation with varying degree of supervision. 

\textbf{Fully-supervised methods} aim to learn a mapping from 2D information to 3D given pairs of 
2D-3D correspondences as supervision. The recent methods in this direction adopt 
deep neural networks to directly predict 3D poses from 
images~\cite{LiC14,Sijin2015iccv,zhou2016deep,tekin2016structured}. 
Training the data hungry neural networks, however, requires large amounts
of training images with accurate 3D pose annotations which are very hard to acquire, in particular, in unconstrained scenarios.  
To this end, the approaches in~\cite{rogez2016mocap,chen20163dv,varol17b} try to 
augment the training data using synthetic images, however, still need real data 
to obtain good performance. More recent methods try to improve the performance by
incorporating additional data with weak supervision \ie, 2D pose annotations~\cite{popa2017CVPRmultitask,pavlakos2017volumetric,sun2017compositional,zhou2017weakly, dabral18SFM, sun18integeral}, boolean geometric relationship between body parts~\cite{pons2014posebits, pavlakos2018ordinal, RonchiAEP18}, action labels~\cite{luvizon20182d}, and temporal consistency~\cite{arnab2019temporal}. 
Adverserial losses during training~\cite{yang20183d} or testing~\cite{tung2017adverserial} have also been used to improve the performance of models trained on fully-supervised data. 

 Other methods alleviate the need of 3D image annotations by directly lifting 2D poses to 3D without using any image information \eg, by learning a regression network from 2D joints to 3D~\cite{martinez2017simple, Moreno_arxiv2016,hossain2018exploiting} or by searching nearest 3D poses in large databases using 2D projections as the query~\cite{pons2014posebits,chen2017matching,iqabl2018dual}. Since these methods do not use image information for 3D pose estimation, they are prone to re-projection ambiguities and can also have discrepancies between the 2D and 3D poses. 

In contrast, in this work, we present a method that combines the benefits of both paradigms \ie, it estimates 3D pose from an image input, hence, can  handle the re-projection ambiguities, but does not require any images with 3D pose annotations. 

\textbf{Semi-supervised methods} require only a small subset of training data with 3D annotations and assume no or weak supervision for the rest. The approaches~\cite{rohdin2018geometry,rohdin2018multiview, yao2019monet} assume that multiple views of the same 2D pose are available and use multi-view  constraints for supervision. 
Closest to our approach in this category is~\cite{rohdin2018multiview} in that it also uses multi-view consistency to supervise the pose estimation model. However, their method is prone to degenerated solutions and its solution space cannot be constrained easily. Consequently, the requirement of images with 3D annotations is inevitable for their approach. In contrast, our method is weakly-supervised. We constrain the solution space of our method such that the 3D poses can be learned without any 3D annotations. In contrast to~\cite{rohdin2018multiview}, our approach can easily be applied to in-the-wild scenarios as we will show in our experiments. 
The approaches~\cite{tung2017self,wu2016eccv} use 2D pose annotations and re-projection losses to improve the performance of models pre-trained using synthetic data. 
In~\cite{li2019boosting}, a pre-trained model is iteratively improved by refining its predictions using temporal information and then using them as supervision for next steps. The approach in~\cite{pavllo2019temporal} estimates the 3D poses using a sequence of 2D poses as input and uses a re-projection loss accumulated over the entire sequence for supervision. While all of these methods demonstrate impressive results, their main limiting factor is the need of ground-truth 3D data.

\textbf{Weakly-supervised methods} do not require paired 2D-3D data and only use weak supervision in form of motion-capture data~\cite{tome2017lifting}, images/videos with 2D annotations~\cite{novotny2019C3DPO, wang2019nrsfm}, collection of 2D poses~\cite{drove2018can3d, wandt2019repnet, chen2019unsupervised}, or multi-view images~\cite{pavlakos2017harvesting, kocabas2019epipolar}. Our approach also lies in this paradigm and learns to estimate 3D poses from unlabeled multi-view data. In~\cite{tome2017lifting}, a probabilistic 3D pose model learned using motion-capture data is integrated into a multi-staged 2D pose estimation model to iteratively refine 2D and 3D pose predictions. The approach~\cite{novotny2019C3DPO} uses a re-projection loss to train the pose estimation model using images with only 2D pose annotations. Since re-projection loss alone is insufficient for training, they factorize the problem into the estimation of view-point and shape parameters and provide inductive bias via a canonicalization loss. Similar in spirit, the approaches~\cite{drove2018can3d, wandt2019repnet, chen2019unsupervised} use collection of 2D poses with re-projection loss for training and use adversarial losses to distinguish between plausible and in-plausible poses. In~\cite{wang2019nrsfm}, non-rigid structure from motion is used to learn a 3D pose estimator from videos with 2D pose annotations. The closest to our work are the approaches of~\cite{pavlakos2017harvesting, kocabas2019epipolar} in that they also use unlabeled multi-view data for training. The approach of~\cite{pavlakos2017harvesting}, however, requires calibrated camera views that are very hard to acquire in unconstrained environments. 
The approach~\cite{kocabas2019epipolar} estimates 2D poses from multi-view images and reconstructs corresponding 3D pose using Epipolar geometry. The reconstructed poses are then used for training in a fully-supervised way. The main drawback of this method is that the 3D poses remain fixed throughout the training, and the errors in 3D reconstruction directly propagate to the trained models. This is, particularly, problematic if the multi-view data is captured in challenging outdoor environments where 2D pose estimation may fail easily. In contrast, in this work, we propose an end-to-end learning framework which is robust to challenges posed by the data captured in in-the-wild scenarios. It is trained using a novel objective function which can simultaneously optimize for 2D and 3D poses. In contrast to~\cite{kocabas2019epipolar},  our approach can also improve 2D predictions using unlabeled multi-view data. We evaluate our approach on two challenging datasets where it outperforms existing methods for semi-/weakly-supervised learning by a large margin. 
\section{Method}
Our goal is to train a convolutional neural network $\mathcal{F}(\bf{I},\theta)$ parameterized by weights $\theta$ that, given an RGB image $\bf{I}$ as input, estimates the 3D body pose $\mathbf{P}={\{\mathbf{p}_j\}}_{j \in J}$ consisting of 3D locations $\mathbf{p}_j=(x_j,y_j,z_j) \in \mathbb{R}^3$ of $J$ body joints with respect to the camera. 

We do not assume any training data with paired 2D-3D annotations and learn the parameters $\theta$ of the network in a weakly-supervised way using unlabeled multi-view images and an independent collection of images with 2D pose annotations. 
To this end, we build on the 2.5D pose representation of~\cite{iqbal2018hand} for hand pose estimation and extend it to human body. This 2.5D pose representation has several key features that allow us to exploit multi-view information and devise loss functions for weakly-supervised training.

In the following, we first recap the 2.5D pose representation (Sec.~\ref{sec:2.5D_pose}) and the approach to reconstruct absolute 3D pose from it (Sec.~\ref{sec:3d_reconstruction}). We then describe a fully-supervised approach to regress the 2.5D pose using a convolutional neural network~(Sec.\ref{sec:2.5D_pose_regression}) followed by our proposed method for weakly-supervised training in Sec.~\ref{sec:weakly-supervised-training}. 
An overview of the proposed approach can be seen in Fig.~\ref{fig:overview}.

\subsection{2.5D Pose Representation}
\label{sec:2.5D_pose}
Many existing methods~\cite{sun2017compositional,sun18integeral,pavlakos2018ordinal} for 3D body pose estimation adopt a 2.5D pose representation $\mathbf{P}^{\mathrm{2.5D}}=\{\mathbf{p}^{\mathrm{2.5D}}_j=(u_j, v_j,z^r_j)\}_{j \in J}$ where  $u_j$ and $v_j$ are the 2D projection of the body joint $j$ on a camera plane and $z^r_j = z_\mathrm{root} - z_j$ represents its metric depth with respect to the root joint. This decomposition of 3D joint locations into their 2D projection and relative depth has the advantage that additional supervision from in-the-wild images with only 2D pose annotations can be used  for better generalization of the trained models. However, this representation does not account for scale ambiguity present in the image which might lead to ambiguities in predictions.

The 2.5D representation of~\cite{iqbal2018hand}, however, differs from the rest in terms of scale normalization of 3D poses. Specifically, they scale normalize the 3D pose $\mathbf{P}$ such that a specific pair of body joints has a unit distance:
\begin{equation}
\label{eqt:normalization}
\mathbf{\hat{P}} = \dfrac{\mathbf{P}}{s},
\end{equation}
where $s=\Vert{\mathbf{p}_k - \mathbf{p}_l}\Vert_2$ is estimated independently for each pose. The pair $(k,l)$ corresponds to the indices of the joints used for scale normalization. The resulting scale normalized 2.5D pose representation
$\hat{\mathbf{p}}^{\mathrm{2.5D}}_j=(u_j,v_j,\hat{z}^r_j)$ is agnostic to the scale of the person. This not only makes it easier to be estimated from cropped RGB images, but also allows to reconstruct the absolute 3D pose of the person up to a scaling factor in a fully differentiable manner as described next. 

\subsubsection{Differentiable 3D Reconstruction}
\label{sec:3d_reconstruction}
Given the 2.5D pose $\hat{\mathbf{P}}^{\mathrm{2.5D}}$, we need to find the depth $\hat{z}_\mathrm{root}$  of the root joint to reconstruct the scale normalized 3D locations $\hat{\mathbf{P}}$ of body joints using perspective projection:
\begin{equation}
\label{eqt:perspective}
    \hat{\mathbf{p}}_j = \hat{z}_j\mathbf{K}^{-1}\begin{bmatrix} u_j \\ v_j \\ 1 \end{bmatrix} 
    = (\hat{z}_\mathrm{root}+\hat{z}^r_j)\mathbf{K}^{-1}\begin{bmatrix} u_j \\ v_j \\ 1 \end{bmatrix}.
\end{equation}
The value of $\hat{z}_\mathrm{root}$ can be calculated via the scale normalization constraint:
\begin{equation}
\label{eqt:normalization_constraint}
 (\hat{x}_k - \hat{x}_l)^2 + (\hat{y}_k - \hat{y}_l)^2 + (\hat{z}_k - \hat{z}_l)^2 = 1,
\end{equation}
which leads to an analytical solution as derived in~\cite{iqbal2018hand}. 
Since all operations for 3D reconstruction are differentiable, we can devise loss functions that directly operate on the reconstructed 3D poses.

In the rest of this paper, we will use the scale normalized 2.5D pose representation. We use the distance between the neck and pelvis joints to calculate the scaling factor $s$. 

\subsection{2.5D Pose Regression}
\label{sec:2.5D_pose_regression}
Since the 3D pose can be reconstructed analytically from 2.5D pose, we train the network to predict 2.5D pose and implement 3D reconstruction as an additional parameter-free layer. To this end, we adopt the 2.5D heatmap regression approach of~\cite{iqbal2018hand}. Specifically, given an RGB image as input, the network produces $2J$ channels as output with $J$ channels for 2D heatmaps ($\mathbf{H}^{\mathrm{2D}}$) while the remaining $J$ channels are regarded as latent depth maps  $\mathbf{H}^{z}$. The 2D heatmaps are converted to 2D pose coordinates $(u_j, v_j)$ by first normalizing them using spatial $\mathrm{softmax}$, \ie, $\bar{\mathbf{H}}^{\mathrm{2D}}_j = \mathrm{softmax}(\mathbf{H}^{\mathrm{2D}}_j, \lambda)$, 
and then using the $\mathrm{soft{\mhyphen}argmax}$ operation:
\begin{equation}
\label{eqt:softargmax}
\small
u_j = \sum_{u, v \in \mathcal{U}} u\cdot\bar{\mathbf{H}}^{\mathrm{2D}}_j(u,v);\quad 
v_j = \sum_{u, v \in \mathcal{U}} v\cdot\bar{\mathbf{H}}^{\mathrm{2D}}_j(u,v),
\end{equation}
where $\mathcal{U}$ is a 2D grid sampled according to the effective stride size of the network, and $\lambda$ is a constant that controls the temperature of the normalized heatmaps. 

The relative scale normalized depth value $\hat{z}^r_j$ for each joint can then be obtained as the summation of the element-wise multiplication of $\bar{\mathbf{H}}^{\mathrm{2D}}_j$ and latent depth maps $\mathbf{H}^{z}_{j} $:
\begin{equation}
\hat{z}^r_j = \sum_{u,v} \bar{\mathbf{H}}^{\mathrm{2D}}_j \odot  \mathbf{H}^{z}_j.
\end{equation}
Given the 2D pose coordinates $\{(u_j, v_j)\}_{j \in J}$, relative depths $\hat{\mathbf{z}}^r=\{\hat{z}_j^r\}_{j\in J}$ and intrinsic camera parameters $\mathbf{K}$, the 3D pose can be reconstructed as explained in Sec.~\ref{sec:3d_reconstruction}. 

In the fully-supervised (FS) setting, the network can be trained using the following loss function:
\begin{equation}
\label{eqt:loss_fs}
\mathcal{L}_\text{FS} = \mathcal{L}_\text{H}(\mathbf{H}^{\mathrm{2D}}, {\mathbf{H}}^{\mathrm{2D}}_\text{gt}) + \psi \mathcal{L}_{\mathbf{z}}(\hat{\mathbf{z}}^r, \hat{\mathbf{z}}^r_\text{gt}),
\end{equation}
where ${\mathbf{H}}^{\mathrm{2D}}_\text{gt}$ and $\hat{\mathbf{z}}^r_\text{gt}$ are the ground-truth 2D heatmaps and ground-truth scale-normalized relative depth values, respectively. We use mean squared error as the loss functions $\mathcal{L}_{\text{H}}(\cdot)$ and $\mathcal{L}_{\mathbf{z}}(\cdot)$. 

We make one modification to the original loss to better learn the confidence scores of predictions. Specifically, in contrast to~\cite{iqbal2018hand}, we do not learn 2D heatmaps in a latent way. Instead, we chose to explicitly supervise the 2D heatmap predictions via ground-truth heatmaps with Gaussian distributions at the true joint locations. We will rely on the confidence scores to devise a weakly-supervised loss that is robust to uncertainties in 2D pose estimates, as described in the following section. 

\subsection{Weakly-Supervised Training}
\label{sec:weakly-supervised-training}
We describe our proposed approach for training the regression network in a weakly-supervised way without any 3D annotations. For training, we assume a set $\mathbf{M} = \{\{\mathbf{I}_c^n\}_{c \in C_n}\}_{n \in N}$ of $N$ samples, with the $n^\text{th}$ sample consisting of $C_n$ camera views of a person in same body pose. The multi-view images can be taken at the same time using multiple cameras, or using a single camera assuming a static body pose over time. We do not assume knowledge of extrinsic camera parameters. Additionally, we use an independent set of images annotated only with 2D poses which is available abundantly or can be annotated by people even for in-the-wild data. 
For training, we optimize the following weakly-supervised (WS) loss function:
\begin{equation}
\label{eqt:loss_ws}
\mathcal{L}_\text{WS} = \mathcal{L}_\text{H}(\mathbf{H}^{\mathrm{2D}}, {\mathbf{H}}^{\mathrm{2D}}_\text{gt}) + \alpha \mathcal{L}_\text{MC}(\mathbf{M}) + \beta \mathcal{L}_\text{B}(\hat{\mathbf{L}}, \hat{\mu}^L),
\end{equation}
where $\mathcal{L}_\text{H}$ is the 2D heatmap loss, $\mathcal{L}_\text{MC}$ is the multi-view consistency loss, and $\mathcal{L}_\text{B}$ is the limb length loss. 

Recall that, given an RGB image, our goal is to estimate the scale normalized 2.5D pose $\hat{\mathbf{P}}^{\mathrm{2.5D}}=\{\mathbf{p}^{\mathrm{2.5D}}_j=(u_j, v_j,\hat{z}^r_j)\}_{j \in J}$ from which we can reconstruct the scale normalized 3D pose $\hat{\mathbf{P}}$ as explained in Sec.~\ref{sec:3d_reconstruction}. While $\mathcal{L}_\text{H}$ provides supervision for 2D pose estimation, the loss $\mathcal{L}_\text{MC}$ supervises the relative depth component ($\hat{\mathbf{z}}^r$). The limb length loss $\mathcal{L}_\text{B}$ further ensures that the reconstructed 3D pose $\hat{\mathbf{P}}$ has plausible limb lengths. In the following, we explain these loss functions in more detail. 

\textbf{Heatmap Loss} ($\mathcal{L}_\text{H}$) measures the difference between the predicted 2D heatmaps $\mathbf{H}^{\mathrm{2D}}$ and ground-truth heatmaps $\mathbf{H}^{\mathrm{2D}}_\text{gt}$ with Gaussian distribution at the true joint location. It operates only on images annotated with 2D poses and is assumed to be zero for all other images.

\textbf{Multi-View Consistency Loss} ($\mathcal{L}_\text{MC}$) enforces that the 3D pose estimates obtained from different views should be identical up to a rigid transform. Formally, given a multi-view training sample $\mathbf{M} = \{\mathbf{I}_c\}_{c \in C}$ with $C$ camera views, we define the multi-view consistency loss as the weighted sum of the difference between the 3D joint locations  across different views after the rigid alignment:
\begin{equation}
\label{eqt:loss_mc}
\mathcal{L}_\text{MC} = \sum_{\substack{c,c' \in C \\ c \neq c'}}  \sum_{j \in J} \phi_{j,c}\phi_{j,c'} \cdot d(\hat{\mathbf{p}}_{j,c}, \mathbf{R}^{c'}_{c}\hat{\mathbf{p}}_{j,c'}),
\end{equation}
where
\begin{equation}
\phi_{j,c}=\mathbf{H}^{\mathrm{2D}}_{j,c}(u_{j,c}, v_{j,c})~~\text{and}~~\phi_{j,c'}=\mathbf{H}^{\mathrm{2D}}_{j,c'}(u_{j,c'}, v_{j,c'}) \nonumber
\end{equation}
are the confidence scores of the $j^\text{th}$ joint in camera viewpoint $\mathbf{I}_c$ and $\mathbf{I}_{c'}$, respectively. The $\hat{\mathbf{p}}_{j,c}$ and $\hat{\mathbf{p}}_{j,c'}$ are the scale normalized 3D coordinates of the $j^\text{th}$ joint estimated from viewpoint $\mathbf{I}_c$ and $\mathbf{I}_{c'}$, respectively. $\mathbf{R}^{c'}_{c} \in \mathbb{R}^{3\times4}$ is a rigid transformation matrix that best aligns the two 3D poses, and $d$ is the distance metric used to measure the difference between the aligned poses. In this work, we use $L_1$-norm as the distance metric $d$. In order to understand the contribution of $\mathcal{L}_\text{MC}$ more clearly, we can rewrite the distance term in~\eqref{eqt:loss_mc} in terms of the 2.5D pose representation  using~\eqref{eqt:perspective}, \ie:
\begin{align}
\label{eqt:loss_mc_2.5D}
d(\hat{\mathbf{p}}_{j,c}, \mathbf{R}^{c'}_{c}\hat{\mathbf{p}}_{j,c'}) = \nonumber\\
d((\hat{z}_{\mathrm{root},c}\!\!+\!\hat{z}^r_{j,c})\mathbf{K}^{-1}_c&\!\!\!\begin{bmatrix} u_{j,c} \\ v_{j,c} \\ 1 \end{bmatrix}\!\!\!,
 \mathbf{R}^{c'}_{c}\!(\hat{z}_{\mathrm{root},c'}\!\!+\!\hat{z}^r_{j,c'})\mathbf{K}^{-1}_{c'}\!\!\!\begin{bmatrix} u_{j,c'}\\ v_{j,c'} \\ 1 \end{bmatrix}\!).
\end{align}

Let us assume that the 2D coordinates $(u_{j,c},v_{j,c})$ and $(u_{j,c'},v_{j,c'})$ are predicted accurately due to the loss $\mathcal{L}_\text{H}$ and the camera \textit{intrinsics} $\mathbf{K}_{c}$ and $\mathbf{K}_{c'}$ are known. For simplicity, let us also assume the ground-truth transformation $\mathbf{R}^{c'}_{c}$ between the two views is known. Then, the only way for the network to minimize the difference $d(.,.)$ is to predict the correct values for relative depths $\hat{z}^r_{j,c}$ and $\hat{z}^r_{j,c'}$. Hence, the joint optimization of the losses $\mathcal{L}_\text{H}$ and $\mathcal{L}_\text{MC}$ allows us to learn correct 3D poses using only weak supervision in form of multi-view images and 2D pose annotations. Without the loss $\mathcal{L}_\text{H}$ the model can lead to degenerated solutions. 

While in many practical scenarios, the transformation matrix $\mathbf{R}^{c'}_{c}$ can be known a priori via extrinsic calibration, we, however, assume it is not available and estimate it using predicted 3D poses and Procrustes analysis as follows:
\begin{equation}
\label{eqt:transformation_matrix}
\mathbf{R}_{c}^{c'} = \argmin_\mathbf{R} \sum_{j \in J} \phi_{j,c}\phi_{j,c'} \Vert \hat{\mathbf{p}}_{j,c} - \mathbf{R} \hat{\mathbf{p}}_{j,c'} \Vert^2_2. 
\end{equation}
During training, we follow~\cite{rohdin2018multiview} and do not back-propagate through the optimization of transformation matrix~\eqref{eqt:transformation_matrix}, since it leads to numerical instabilities arising due to singular value decomposition. Note that the gradients from $\mathcal{L}_\text{MC}$ not only influence the depth estimates, but also affect heatmap predictions due to the calculation of $\hat{z}_{\mathrm{root}}$ in~\eqref{eqt:normalization_constraint}. Therefore, $\mathcal{L}_\text{MC}$ can also fix the errors in 2D pose estimates as we will show in our experiments.

\textbf{Limb Length Loss} ($\mathcal{L}_\text{B}$) measures the deviation of the limb lengths of predicted 3D pose from the mean bone lengths:
\begin{equation}
\mathcal{L}_{\text{B}} =  \sum_{j,j' \in \mathcal{E}} \phi_{j} \phi_{j'} (\Vert \hat{\mathbf{p}}_j - \hat{\mathbf{p}}_{j'} \Vert - \hat{\mu}^{L}_{j,j'})^2,
\end{equation}
where $\mathcal{E}$ corresponds to the used kinematic structure of the human body and $\hat{\mu}^{L}_{j,j'}$ is the scale normalized mean limb length for joint pair $(j,j')$. 
Since the limb lengths of all people will be roughly the same after scale normalization~\eqref{eqt:normalization}, this loss ensures that the predicted poses have plausible limb lengths. During training, we found that having a limb length loss leads to faster convergence. 

\textbf{Additional Regularization} We found that if a large number of samples in multi-view data have a constant background, the network learns to recognize these images and starts predicting same 2D pose and relative depth values for such images. Interestingly, it predicts correct values for other samples. In order to prevent this, we incorporate an additional regularization loss for such samples. Specifically, we run a pre-trained 2D pose estimation model and generate pseudo ground-truths by selecting joint estimates with confidence score greater than a threshold $\tau = 0.5$. These pseudo ground-truths are then used to enforce the 2D heatmap loss $\mathcal{L}_\text{H}$, which prevents the model from predicting degenerated solutions. We generate the pseudo ground-truths once at the beginning of the training and keep them fixed throughout. Specifically, we use the regularization loss for images from Human3.6M~\cite{h36m_pami} and MPII-INF-3DHP~\cite{mono20173dhp}  that are both recorded in controlled indoor settings. While the regularization may reduce the impact of $\mathcal{L}_\text{MC}$ on 2D poses, the gradients from $\mathcal{L}_\text{MC}$ will still influence the heatmap predictions of body joints that were not detected with high confidence (see Fig.~\ref{fig:impact_of_mqc}). 
\section{Experiments}

We evaluate our proposed approach for weakly-supervised 3D body pose learning and compare it with the state-of-the-art methods.
Additional training and implementation details can be found in Appendix~\ref{sec:implementation_details}.

\subsection{Datasets}
We use two large-scale datasets, Human3.6M~\cite{h36m_pami} and MPII-INF-3DHP~\cite{mono20173dhp} for evaluation. For weakly-supervised training, we also use the MannequinChallenge dataset~\cite{li2019mannequin} 
and 
MPII Human Pose dataset~\cite{andriluka14cvpr}. The details of each dataset are as follows.

\noindent\textbf{Human3.6M (H36M)}~\cite{h36m_pami} 
provides images of actors performing a variety of actions from four views. 
We follow the standard protocol and use five subjects (S1, S5, S6, S7, S8) for training and test on two subjects (S9 and S11).

\noindent\textbf{MPII-INF-3DH (3DHP)}~\cite{mono20173dhp} provides ground-truth 3D poses obtained using markerless motion-capture system.
Following the standard protocol~\cite{mono20173dhp}, we use five chest height cameras for training. The test-set consists of six sequences with actors performing a variety activities. 

\noindent\textbf{MannequinChallenge Dataset (MQC) }~\cite{li2019mannequin} provides in-the-wild videos of people in static poses while a hand-held camera pans around the scene. The videos do not come with any ground-truth annotations, however, the data is very adequate for our proposed weakly-supervised approach using multi-view consistency. The dataset consists of three splits for training, validation and testing. In this paper, we use ${\sim}3300$ videos from training and validation set as proposed by~\cite{li2019mannequin}, but in practice one could download an immense amount of such videos from YouTube (\#MannequinChallenge). We will show in our experiments that using these in-the-wild videos during training yields better generalization,
in particular, when there is a significant domain gap between the training and testing set. Since the videos can have multiple people inside each frame, they have to be associated across frames to obtain the required multi-view data. To this end, we adopt the pose based tracking approach of~\cite{FlowTrack} and generate person tracklets from each video. For pose estimation, we use a HRNet-w32~\cite{SunXLW19} model pretrained on MPII Pose dataset~\cite{andriluka14cvpr}. In order to avoid training on noisy data, we discard significantly occluded or truncated people. We do this by  discarding all poses that have more than half of the estimated body joints with confidence score lower than a threshold $\tau{=}0.5$. We also discard poses in which neck or pelvis joints have confidence lower than $\tau{=}0.5$ since both joints are important for $\hat{z}_{\mathrm{root}}$ reconstruction using~\eqref{eqt:normalization_constraint}. Finally, we discard all tracklets with the length lower than 5 frames. This gives us 11,413 multi-view tracklets with 241k images in total. The minimum and maximum length of the tracklets is 5 and 140 frames, respectively. 

\noindent\textbf{MPII Pose Dataset (MPII)~\cite{andriluka14cvpr}} provides 2D pose annotations for 28k in-the-wild images. 

\subsection{Evaluation Metrics}
For evaluation on H36M, we follow the standard protocols and use MPJPE (Mean Per Joint Position Error), N-MPJPE (Normalized-MPJPE) and  P-MPJPE (Procrustes-aligned MPJPE) for evaluations. MPJPE measures the mean euclidean error between the ground-truth and estimated location of 3D joints after root alignment. While NMPJPE~\cite{rohdin2018multiview} also aligns the scale of the predictions with ground-truths, PMPJPE aligns both the scale and rotations using Procrustes analysis. For evaluation on 3DHP dataset, we follow~\cite{mono20173dhp} and also report PCK (Percentage of Correct
Keypoints) and Normalized-PCK as defined in~\cite{rohdin2018multiview}. PCK measures the percentage of predicted body joints that lie within the radius of 150mm from their ground-truths. 3DHP evaluation protocol uses 14 joints for evaluation excluding the pelvis joint which is used for alignment of the poses.  

\subsection{Ablation Studies}

\begin{table}[t]
\centering
\small
\begin{tabularx}{1\columnwidth}{X|ccc|c|c}
\toprule
 \multirow{3}{3cm}{Method} &  \multicolumn{3}{c|}{Supervision} & \multicolumn{2}{c}{Error}  \\
            & ~~~2D~~~  & 3D    & MV    &  2D  px  & 3D mm \\
\midrule
\midrule
FS          & H+M     & H     & -     &        5.9  & 55.5  \\ 
\midrule
WS +  $\mathbf{R}$    & H+M     & -     & H     &       6.1    & 57.2\\ 
WS          & H+M     & -     & H     &       6.1    & 59.3   \\ 
\midrule
2d-only     & M         & -     & -     &       8.9        & - \\ 
WS + $\mathbf{R}$    & M         & -     & H     &       8.3        & 62.3 \\ 
WS          & M         & -     & H              &       8.4        & 69.1 \\ 
WS                  & M         & -     & I    &      9.0    & 106.2\\ 
WS                  & M         & -     & I+Q    &      9.1    & 93.6 \\ 
WS                   & M         & -    & H+I+Q  &       8.4    & 67.4 \\ 
WS + $\mathbf{R}$    & M         & -    & H+I+Q    &       8.4    & 60.3\\ 
\bottomrule
\end{tabularx}
\vspace{1mm}
\caption{\textbf{Ablative study:} We provide results when different levels of supervision are used to train the proposed weakly-supervised method. FS: Fully-Supervised, WS: Weakly-Supervised, MV: Multi-View, H: H36M, M: MPII, I: 3DHP, Q: MQC. No 3D supervision is used for all experiments except FS. \vspace{-3mm}}
\label{tab:ablation}

\end{table}

\begin{figure*}[t]
    \centering
    \includegraphics[trim={0 0cm 0 0cm},clip,width=\textwidth]{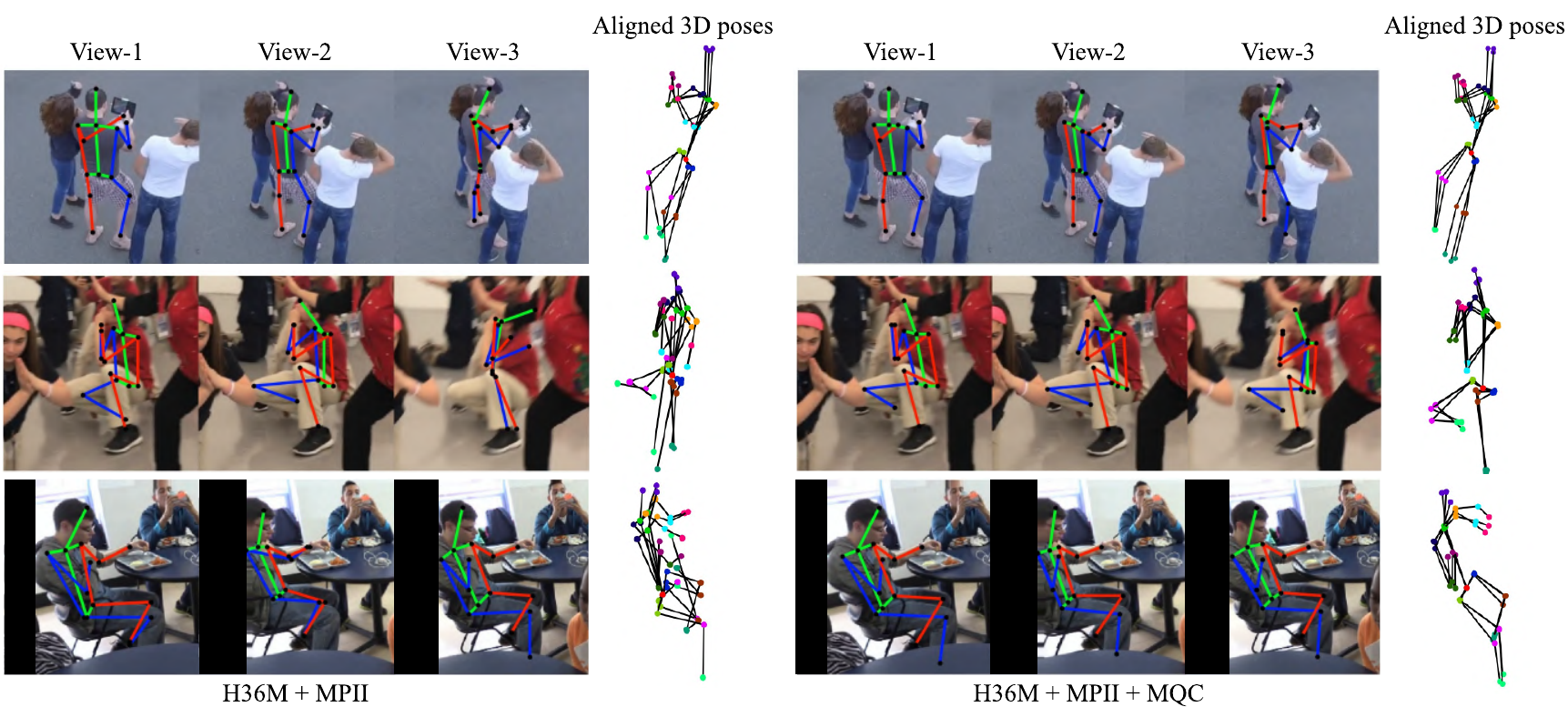}
    \vspace{-5mm}
    \caption{Impact of using MQC dataset. We run the trained models on the tracks taken from MQC dataset and align the estimated 3D poses using~\eqref{eqt:transformation_matrix}. Since people in MQC dataset do not move, the aligned poses should be very similar. Adding MQC dataset for training \textbf{(right)} yields
more consistent 3D pose estimates as compared to when only H36M is used \textbf{(left)} for multi-view consistency loss. Note that our proposed approach can also fix the errors in 2D pose estimates in the unlabeled multi-view data. \vspace{-3mm}}
    \label{fig:impact_of_mqc}
\end{figure*}

Tab.~\ref{tab:ablation} evaluates the impact of different levels of supervision for training with the proposed approach. We use  H36M for evaluation. We start with a fully-supervised setting (FS) which uses 2D supervision from H36M and MPII (2D=H+M) datasets and 3D pose supervision from H36M (3D=H). No multi-view (MV) data is used in this case. The fully-supervised model yields a MPJPE of 5.9px  and 55.5mm for 2D and 3D pose estimation, respectively. We then remove the 3D supervision and instead train the network using the proposed approach for weakly-supervised learning (WS+$\mathbf{R}$). The MV data is taken from H36M (MV=H). For this experiment, we assume that the 2D pose annotations for MV data are available (2D=H+M) and the camera extrinsics $\mathbf{R}$ are known. This setting is essentially similar to fully-supervised case since the 2D poses from different views can be triangulated using the known $\mathbf{R}$. Training the network under this setting, however, serves as a sanity check that the proposed weakly-supervised approach works as intended which can be confirmed by the obtained 3D pose error of 57.2mm.  If $\mathbf{R}$ is unknown (WS) and is obtained from estimated 3D poses using~\eqref{eqt:transformation_matrix}, the error increases slightly to 59.3mm. 

All of the aforementioned settings assume that the MV data is annotated with 2D poses which is infeasible to collect in large numbers. Therefore, we have designed the proposed method to work with MV data without even 2D annotations. Next, we remove the 2D supervision from MV data and only use MPII dataset for 2D supervision (2D=M). For reference, we also report the error of a 2D-only model trained on MPII dataset which yields a 2D pose error of 8.9px. 
Training without 2D pose annotations for MV data with and without ground-truth $\mathbf{R}$ yields errors of 62.3mm (WS+$\mathbf{R}$) and 69.1mm (WS), respectively, as compared to 57.2mm and 59.3mm when the 2D pose annotations are available. While using ground-truth $\mathbf{R}$ always yields better performance, for the sake of easier applicability, in the rest of this paper, we assume it to be unknown unless specified otherwise. It is also interesting to note that the 2D pose error decreases from 8.9px to 8.3px when the multi-view consistency loss~\eqref{eqt:loss_mc} is used. Some qualitative examples of improvements in 2D poses can be seen in~Fig.~\ref{fig:impact_of_mqc}.  

We also evaluate the case when the training data is recorded in different settings than testing data. For this, we use 3DHP for training (MV=I) and test on H36M. Since the images of 3DHP are very different from H36M, it leads to a very high error of 106.2mm. Adding the generated training data from MQC dataset (MV=I+Q) significantly reduces the error to 93.6mm which demonstrates the effectiveness of in-the-wild data from MQC. Combining all three datasets (MV=H+I+Q) reduces the error further to 67.4mm as compared to 69.1mm when only H36M dataset was used for training. We also provide the results when ground-truth $\mathbf{R}$ is known (WS+$\mathbf{R}$) for H36M and 3DHP datasets (MV=H+I+Q) which shows a similar behaviour and decreases the error from 62.3mm to 60.3mm. 

In our experiments, we found that training only on MQC dataset is not sufficient for convergence and it has to be combined with another dataset which provides multi-view data from more distant viewing angles. This is likely because most videos in MQC dataset do not capture same person from very different viewing angles, whereas datasets such as H36M and 3DHP provide images from cameras with sufficiently large baselines.

\subsection{Comparison with the State-of-the-Art}

\begin{table}[t]
\scriptsize
\centering
\begin{tabularx}{1\columnwidth}{X|ccc}
\toprule
\bf Methods & \bf MPJPE\ $\downarrow$ & \bf NMPJPE\ $\downarrow$ & \bf PMPJPE\ $\downarrow$ \\
\midrule
\midrule
\multicolumn{4}{c}{Fully-Supervised Methods}\\
\midrule
Rogez \etal \cite{rogez2017lcr} (CVPR'17) & 87.7 & -  & 71.6  \\
Habibie \etal \cite{habibie2019wild} (ICCV'19) & - & 65.7 & - \\
Rhodin \etal \cite{rohdin2018multiview} (CVPR'18)& 66.8 & 63.3 & 51.6  \\
Zhou \etal \cite{zhou2017weakly} (ICCV'17) & 64.9 & - & - \\
Martinez \etal \cite{martinez2017simple} (ICCV'17) & 62.9 & - & 47.7  \\
Sun \etal~\cite{sun2017compositional} (ICCV'17) \bf* & 59.6 & - & - \\ 
Yang \etal~\cite{yang20183d} (CVPR'18)    & 58.6 & - & - \\
Pavlakos \etal \cite{pavlakos2018ordinal} (CVPR'18)& 56.2 & - & -  \\
Sun \etal~\cite{sun18integeral} (ECCV'18)\bf*  & \bf 49.6 & - & {40.6} \\ 
Kocabas \etal \cite{kocabas2019epipolar} (CVPR'19)\bf* & 51.8 & 51.6 & 45.0  \\
Ours - H - baseline & 55.5 &51.4   &41.5 \\
Ours\textbf{*} - H - baseline & 50.2 & \bf 49.9   & \bf 36.9 \\ 
Ours - H+I+Q & 56.1 & 52.7   & 45.9 \\
\midrule
\midrule
\multicolumn{4}{c}{Semi-Supervised Methods - only Subject-1 is used for training }\\
\midrule
Rohdin \etal \cite{rohdin2018geometry} (ECCV'18)  & 131.7 & 122.6 & 98.2 \\
Pavlakos \etal \cite{pavlakos2019texture} (ICCV'19) & 110.7 & 97.6 & 74.5 \\
Li \etal \cite{li2019boosting} (ICCV'19)          & 88.8  & 80.1  & 66.5 \\                     
Rhodin \etal \cite{rohdin2018multiview} (CVPR'18)  & n/a & 80.1  & 65.1 \\
Kocabas \etal \cite{kocabas2019epipolar} (CVPR'19) & n/a & 67.0 & 60.2 \\
Ours - H & 62.8 & 59.6   &51.4\\ 
Ours - H+I+Q & \bf 59.7 & \bf 56.2   & \bf 50.6 \\
\midrule
\midrule
\multicolumn{4}{c}{Weakly-Supervised Methods - no 3D supervision}\\
\midrule
Pavlakos \etal~\cite{pavlakos2017harvesting} (CVPR'17) & 118.4 & - & - \\
Kanzawa \etal \cite{hmrKanazawa18} (CVPR'18) & 106.8 & - & 67.5 \\
Wandt \etal~\cite{wandt2019repnet} (CVPR'19) & 89.9 & - & - \\
Tome \etal \cite{tome2017lifting} (CVPR'17) & 88.4 & - & -  \\   
Kocabas \etal \cite{kocabas2019epipolar}  (CVPR'19) & n/a & 77.75 & 70.67 \\ 
Chen \etal \cite{chen2019unsupervised} (CVPR'19) & - & - & 68.0 \\
Drover \etal \cite{drove2018can3d} (ECCV-W'18) & - & - & 64.6  \\
Kolotouros \etal \cite{kolotouros2019spin} (ICCV'19) & - & - & 62.0 \\
Wang \etal \cite{wang2019nrsfm} (ICCV'19) & 83.0 & - & 57.5 \\
Ours - H & 69.1 &66.3   &55.9 \\
Ours - H+I+Q & \bf 67.4 & \bf 64.5   & \bf 54.5  \\
\bottomrule
\end{tabularx}
\hspace{5mm}

\caption{ Comparison with the state-of-the-art on H36M dataset. \textbf{*}use ground-truth depth of the root keypoint during inference.\vspace{-3mm}}
\label{table:sota_h36m}
\end{table}

Tab.~\ref{table:sota_h36m} compares the performance of our proposed method with the state-of-the-art on H36M dataset. We group all approaches in three categories; fully-supervised, semi-supervised, and weakly-supervised, and compare the performance of our method under each category.  While fully-supervised methods use complete training set of H36M for 3D supervision, semi-supervised methods use 3D supervision from only one subject (S1) and use other subjects (S5,  S6, S7,  S9) for weak supervision. Weakly-supervised methods do not use any 3D supervision. Some methods also use ground-truth information during inference~\cite{sun2017compositional,sun18integeral,kocabas2019epipolar}. For a fair comparison with those, we also report our performance under the same settings. It is important to note that, many approaches such as~\cite{rohdin2018geometry,rohdin2018multiview,martinez2017simple,rogez2017lcr, yang20183d,chen2019unsupervised,drove2018can3d} estimate root-relative 3D pose. Our approach, on the other hand, estimates absolute 3D poses. While our fully-supervised baseline (Ours-H-baseline) performs better or on-par with the state-of-the-art fully-supervised methods, our proposed approach for weakly-supervised learning   significantly outperforms other methods under both semi- and weakly-supervised categories. 

For a fair comparison with other methods, we report results of our method under two settings: i) using H36M and MPII dataset for training (Ours-H), and ii) with multi-view data from 3DHP and MQC as additional weak supervision (Ours-H+I+Q). In the fully-supervised case, using additional weak supervision slightly worsens the performance (55.5mm vs 56.1mm) which is not surprising on a dataset like H36M which is heavily biased to indoor data and have training and testing images recorded with a same background. Whereas, our approach, in particular the data from MQC, is devised for in-the wild generalization. The importance of additional multi-view data, however, can be seen evidently in the semi-/weakly-supervised settings where it decreases the error from 62.8mm to 59.7mm and from 69.1mm to 67.4mm, respectively.

\begin{table}[t]
\scriptsize
\centering
\begin{tabularx}{1\columnwidth}{X|sstt}
\toprule
 \multirow{1}{3cm}{\bf Methods}             &  \textbf{\footnotesize MPJPE}$\downarrow$    & \textbf{\footnotesize NMPJPE}$\downarrow$ &  \textbf{\footnotesize PCK}$\uparrow$    &  \textbf{\footnotesize NPCK}$\uparrow$ \\
\midrule
\midrule
\multicolumn{5}{c}{Fully-Supervised Methods}\\
\midrule
Mehta \etal~\cite{mehta2017vnect}           & -                 & -                  & 76.6              & - \\
Rohdin \etal~\cite{rohdin2018multiview}     & n/a               & 101.5              & n/a               & 78.8 \\
Kocabas \etal~\cite{kocabas2019epipolar}*   & 109.0             & 106.4              & 77.5              & 78.1 \\
Ours                                        &  \bf110.8 & \bf98.9 & \bf80.2 &  \bf82.3 \\
Ours*                                       &  \bf99.2  & \bf97.2 & \bf83.0 &  \bf83.3 \\
\midrule
\midrule
\multicolumn{5}{c}{Semi-Supervised Methods}\\
\midrule
Rhodin \etal \cite{rohdin2018multiview}     & n/a               & 121.8             & n/a               & 72.7 \\
Kocabas \etal \cite{kocabas2019epipolar}    & n/a               & 119.9             & n/a               & 73.5  \\
Ours                                        & \bf 113.8         & \bf 102.2         & \bf 79.1          & \bf 81.5 \\
\midrule
\midrule
\multicolumn{5}{c}{Weakly-Supervised Methods}\\
\midrule
Kanazwa~\etal~\cite{hmrKanazawa18}          & 169.5             & -                 & 59.6              & - \\
Kolotouros \etal \cite{kolotouros2019spin} & 124.8              & -                 & 66.8              & - \\
Ours                                        & \bf 122.4         & \bf 110.1         & \bf 76.5          & \bf 79.4 \\
\midrule
Kocabas~\textit{et al.}~\cite{kocabas2019epipolar}\textbf{*} + $\mathbf{R}$ & 126.8 & 125.7   & 64.7              & 71.9 \\
Ours\textbf{*} +  $\mathbf{R}$              &  \bf 109.3        & \bf 107.2         & \bf 79.5 & \bf 80.0 \\
\bottomrule
\end{tabularx}
\caption{Comparison with the state-of-the-art on 3DHP dataset. \textbf{*}use ground-truth 3D location of the root joint during inference.\vspace{-3mm}} 
\label{table:sota_3dhp}
\end{table}

Compared to the state-of-the-art method~\cite{kocabas2019epipolar} that also uses multi-view information for weak supervision, our method performs significantly better even though the fully-supervised baselines of both approaches perform similar. This demonstrates the effectiveness of our end-to-end training approach and proposed loss functions that are robust to errors in 2D poses. 
While our weakly-supervised approach does not outperform fully-supervised methods, it performs on-par with many recent fully-supervised approaches.

Tab.~\ref{table:sota_3dhp} compares the performance of our proposed approach with the state-of-the-art on 3DHP dataset. We use our models trained with \textit{Ours-H+I+Q} setting, as described above. We do not use any 3D pose supervision from 3DHP dataset and instead use the same models used for evaluation on H36M dataset. Our proposed approach outperforms all existing methods with large margins under all three categories which also demonstrates the cross dataset generalization of our proposed method.

Some qualitative results of the proposed approach can be seen in the supplementary material.

\section{Conclusion}

We have presented a weakly-supervised approach for 3D human pose estimation in the wild. Our proposed approach does not require any 3D annotations and can learn to estimate 3D poses from unlabeled multi-view data. This is made possible by a novel end-to-end learning framework and a novel objective function which is optimized to predict consistent 3D poses across different camera views. The proposed approach is very practical since the required training data can be collected very easily in in-the-wild scenarios. We demonstrated state-of-the-art performance on two challenging datasets. \\

\noindent\textbf{Acknowledgments:} We are thankful to Kihwan Kim and Adrian Spurr for helpful discussions.

\balance
{\small
\bibliographystyle{ieee_fullname}
\bibliography{pose}
}
\clearpage

\nobalance
\appendix
\section*{Appendix}
We provide implementation details to reproduce the results in the paper. 

\section{Implementation Details}
\label{sec:implementation_details}
We adopt HRNet-w32~\cite{SunXLW19} as the back-bone of our network architecture. We pre-train the model for 2D pose estimation before introducing weakly-supervised losses. This ensures that the 2D pose estimates are sufficiently good to enforce multi-view consistency $\mathcal{L}_\text{MC}$. We use MPII dataset for pre-training. The additional weights for latent depth-maps are  not  pre-trained. We use a maximum of four camera views $C_n{=}4$ to calculate $\mathcal{L}_\text{MC}$. If a sample contains more than four views, we randomly sample four views from it in each epoch. We train the model with a batch size of $256$, where each batch consists of 128 images with 2D pose annotations and 32 unlabeled multi-view samples ($32{\times}4{=}128$ images). For  pre-processing,  we  use  a  person bounding-box to crop the person into a $256\times256$ image such that the person is centered and covers roughly $75\%$ of the image. The training data is augmented by random scaling ($\pm20\%$) and rotation ($\pm30\%$ degrees). We found that the training converges after $60\mathrm{k}$ iterations. The learning rate is set to $5\mathrm{e}{-4}$, which drops to $5\mathrm{e}{-5}$ at $50\mathrm{k}$ iterations following the Adam optimization algorithm. We use $\lambda{=}50$. Since the training objectives~\eqref{eqt:loss_fs} and~\eqref{eqt:loss_ws} consist of multiple loss terms, we balance their contributions by empirically choosing $\psi{=}5$, $\alpha{=}10$, and $\beta{=}100$. 
Since our pose estimation model estimates absolute 3D pose up to a scaling factor, during inference, we approximate the scale using mean bone-lengths from the training data:
\begin{equation}
\label{eqt:scale_recovery}
\hat{s} = \underset{s} {\argmin} \sum_{j,j' \in \mathcal{E}} (s \cdot \Vert \hat{\mathbf{p}}_j - \hat{\mathbf{p}}_{j'} \Vert - \mu^L_{j,j'})^2,
\end{equation}
where is $\mu^L_{j,j'}$ is the mean length of the limb formed by joint pair $(j,j')$. 
In all of our experiments, we use mean lengths from the training set of H36M dataset. 

\end{document}